\pdfoutput=1
\documentclass[10pt,twocolumn,letterpaper]{article}

\usepackage[final]{cvpr}      

\usepackage{graphicx}
\usepackage{subfloat} 
\usepackage{amsmath}
\usepackage{amssymb}
\usepackage{booktabs}
\usepackage{subcaption}
\usepackage{multirow}
\usepackage{bm}
\usepackage{ifthen}
\usepackage{siunitx}
\usepackage{marvosym}
\usepackage{enumerate}
\usepackage{bbm}
\usepackage{colortbl}  
\usepackage{xcolor}
\usepackage[accsupp]{axessibility}  

\usepackage[pagebackref,breaklinks,colorlinks]{hyperref}

\usepackage[capitalize]{cleveref}
\crefname{section}{Sec.}{Secs.}
\Crefname{section}{Section}{Sections}
\Crefname{table}{Table}{Tables}
\crefname{table}{Tab.}{Tabs.}

\def\cvprPaperID{2446} 
\def\confName{CVPR}
\def\confYear{2023}

\begin{document}
\title{CABM: Content-Aware Bit Mapping for Single Image Super-Resolution Network with Large Input}

\author{
Senmao Tian\textsuperscript{1,2}\qquad
Ming Lu\textsuperscript{3}\qquad
Jiaming Liu\textsuperscript{2,4}\qquad
Yandong Guo\textsuperscript{5}\\
Yurong Chen\textsuperscript{3}\qquad
Shunli Zhang\textsuperscript{1}\footnotemark[1]\thanks{This work was supported by the Fundamental Research Funds for the Central Universities (2022JBMC013), the National Natural Science Foundation of China (61976017 and 61601021), and the Beijing Natural Science Foundation (4202056). Shunli Zhang is the corresponding author.}\\
\textsuperscript{1}Beijing Jiaotong University\qquad
\textsuperscript{2}OPPO Research Institute\qquad
\textsuperscript{3}Intel Labs China\qquad\\
\textsuperscript{4}Peking University\qquad
\textsuperscript{5}Beijing University of Posts and Telecommunications\\
{\tt\small \{smtian1204, lu199192\}@gmail.com}\quad{\tt\small slzhang@bjtu.edu.cn}
}
\maketitle

\begin{abstract}
With the development of high-definition display devices, the practical scenario of Super-Resolution (SR) usually needs to super-resolve large input like 2K to higher resolution (4K/8K). To reduce the computational and memory cost, current methods first split the large input into local patches and then merge the SR patches into the output. These methods adaptively allocate a subnet for each patch. Quantization is a very important technique for network acceleration and has been used to design the subnets. Current methods train an MLP bit selector to determine the propoer bit for each layer. However, they uniformly sample subnets for training, making simple subnets overfitted and complicated subnets underfitted. Therefore, the trained bit selector fails to determine the optimal bit. Apart from this, the introduced bit selector brings additional cost to each layer of the SR network. In this paper, we propose a novel method named Content-Aware Bit Mapping (CABM), which can remove the bit selector without any performance loss. CABM also learns a bit selector for each layer during training. After training, we analyze the relation between the edge information of an input patch and the bit of each layer. We observe that the edge information can be an effective metric for the selected bit. Therefore, we design a strategy to build an Edge-to-Bit lookup table that maps the edge score of a patch to the bit of each layer during inference. The bit configuration of SR network can be determined by the lookup tables of all layers. Our strategy can find better bit configuration, resulting in more efficient mixed precision networks. We conduct detailed experiments to demonstrate the generalization ability of our method. The code will be released.
\vskip -0.4cm
\end{abstract}

\begin{figure*}[t!]
	\centering
	\includegraphics[width=0.9\linewidth]{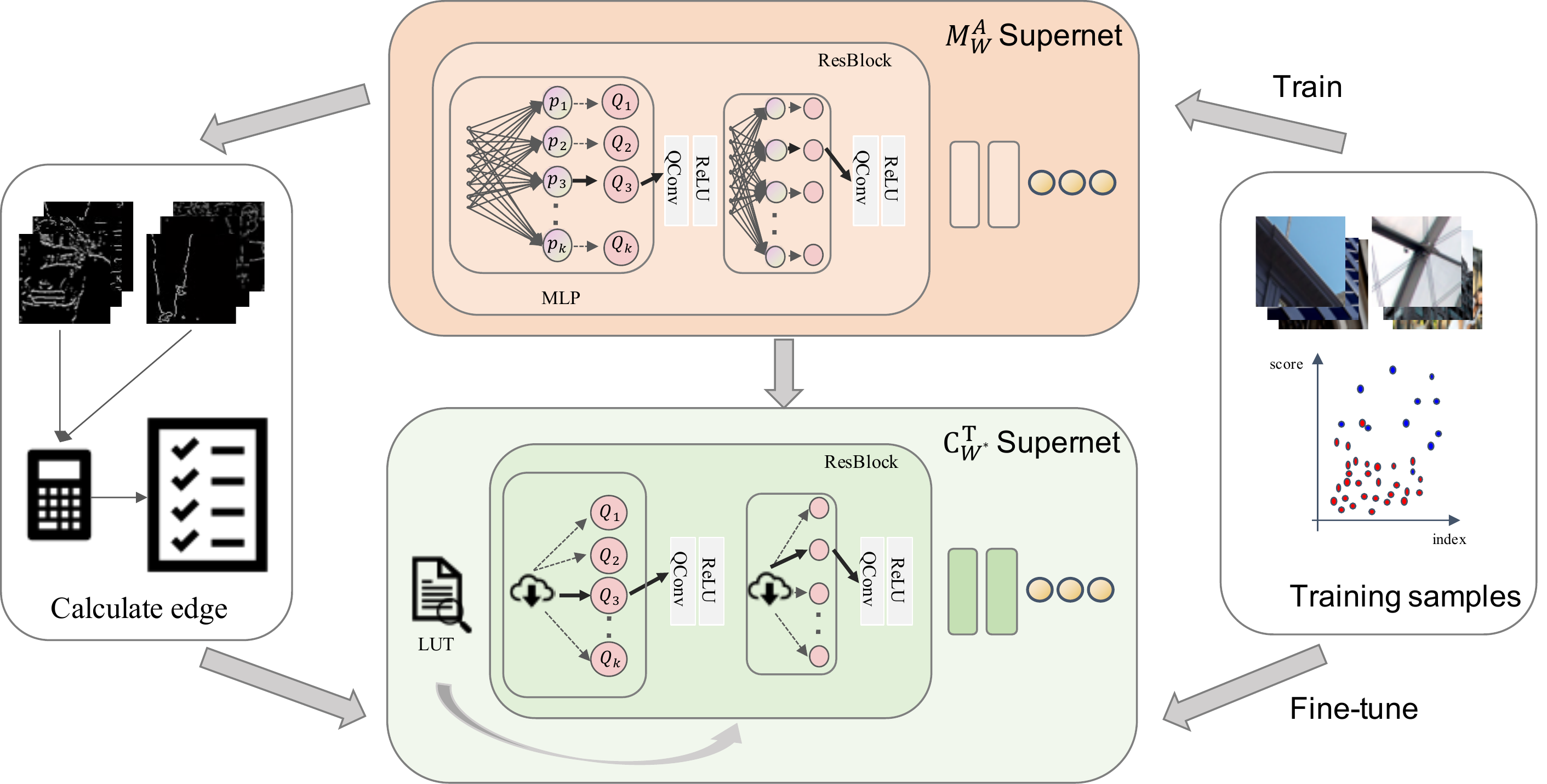} 
	\vskip -0.2cm
	\caption{The pipeline of our CABM method. $p_{i} \in \{p_i\}_{i=1...K}$ is the probability of choosing $i^{th}$ quantization module and each quantization module uses different bit-width to quantize the input activation. During training, our method learns an MLP bit selector to adaptively choose the bit-width for each convolution. While during inference, we use the proposed CABM to build an Edge-to-Bit lookup table to determine the bit-width with negligible additional cost.}
	\label{fig:pipeline}
	\vskip -0.4cm
\end{figure*}

\section{Introduction}
\label{sec:intro}

Single Image Super-Resolution (SISR) is an important computer vision task that reconstructs a High-Resolution (HR) image from a Low-Resolution (LR) image. With the advent of Deep Neural Networks (DNNs), lots of DNN-based SISR methods have been proposed over the past few years \cite{shi2016real,dong2014learning,lim2017enhanced,zhang2018image,kim2016accurate}. While in real-world usages, the resolutions of display devices have already reached 4K or even 8K. Apart from normal 2D images, the resolutions of omnidirectional images might reach even 12K or 16K. Therefore, SR techniques with large input are becoming crucial and have gained increasing attention from the community \cite{kong2021classsr,chen2022arm,wang2022adaptive,hong2022cadyq}.

Since the memory and computational cost will grow quadratically with the input size, existing methods \cite{kong2021classsr,chen2022arm,wang2022adaptive,hong2022cadyq} first split the large input into patches and then merge the SR patches to the output. They reduce the computational cost by allocating simple subnets to those flat regions while using heavy subnets for those detailed regions. Therefore, how to design the subnets is very important for these methods. \cite{kong2021classsr,chen2022arm} empirically decide the optimal channels after lots of experiments to construct the subnets. \cite{wang2022adaptive} proposes to train a regressor to predict the incremental capacity of each layer. Thus they can adaptively construct the subnets by reducing the layers. Compared with pruning the channels or layers, quantization is another promising technique and can achieve more speedup. \cite{hong2022cadyq} trains an MLP bit selector to determine the proper bit for each layer given a patch. However, the introduced MLP of each layer brings additional computational and storage cost. Besides, we observe that \cite{hong2022cadyq} uniformly samples the subnets for training, making simple subnets (low average bit or flat patches) tend to overfit the inputs while complicated subnets (high average bit or detailed patches) tend to underfit the inputs. Therefore, uniform sampling fails to determine the optimal bit for each layer.

To solve the limitations of \cite{hong2022cadyq}, we propose a novel method named Content-Aware Bit Mapping (CABM), which directly uses a lookup table to generate the bit of each layer during inference. However, building a lookup table is difficult since there are thousands of patches and corresponding select bits. We observe that edge information can be an effective metric for patch representation. Therefore, we analyze the relation between the edge information of a patch and the bit of each layer. Inspired by the fact that a MLP selector learns the nonlinear mapping between a patch and the bit, instead of building the Edge-to-Bit lookup table based on linear mapping, we design a tactful calibration strategy to map the edge score of a patch to the bit of each layer. The bit configuration of SR network can be determined by the lookup tables of all layers. Our CABM can achieve the same performance compared with the MLP selectors while resulting in a lower average bit and negligible additional computational cost. Our contributions can be concluded as follows:

\begin{itemize}
	\item We propose a novel method that maps edge information to bit configuration of SR networks, significantly reducing the memory and computational cost of bit selectors without performance loss.
	
	\item We present a tactful calibration strategy to build the Edge-to-Bit lookup tables, resulting in a lower average bit for SR networks.
	
	\item We conduct detailed experiments to demonstrate the generalization ability of our method based on various SR architectures and scaling factors.
\end{itemize}

\section{Related work}

{\bf DNN-based Image Super-Resolution} With the rapid development of DNNs, lots of DNN-based SISR methods have been proposed over the past few years. SRCNN \cite{dong2014learning} is the pioneering work that applies DNNs to the SISR task. Their network consists of three modules including feature extraction, non-linear mapping, and image reconstruction. The following works mostly follow the network design of SRCNN and improve the performance of SISR. For instance, VDSR \cite{kim2016accurate} proposes to use a very deep neural network to predict the residual instead of the HR image. SRResNet \cite{ledig2017photo} introduces the residual block proposed by ResNet \cite{he2016deep} to SR network and improves the performance. EDSR \cite{lim2017enhanced} finds that the BN layer will impair the SR performance and removes it from the structure of SRResNet, further boosting the SR performance. RCAN \cite{zhang2018image} uses the attention mechanism and constructs deeper networks for SR. Real-ESRGAN \cite{wang2021real} extends the powerful ESRGAN \cite{wang2018esrgan} to real-world blind SISR. They introduce a high-order degradation modeling process to simulate complex real-world degradations. USRNet \cite{zhang2020deep} proposes an end-to-end trainable unfolding network that leverages both learning-based methods and model-based methods. Therefore, they can handle the SISR problem with different scaling factors, blur kernels, and noise levels under a unified framework. SwinIR \cite{liang2021swinir} is a strong baseline model that introduces Swin Transformer \cite{liu2021swin} to image restoration. Their non-linear mapping module is composed of several residual Swin Transformer blocks. To reduce the computational cost, there are also many efficient SISR methods. For example, ESPCN \cite{shi2016real} invents the pixel-shuffle layer to obtain the HR output given the LR input. LAPAR \cite{li2020lapar} presents a method based on a linearly-assembled adaptive regression network. Restormer \cite{zamir2022restormer} proposes an efficient Transformer model by making several key designs in the building blocks. All of those methods train one the SR model on large-scale image datasets such as DIV2K \cite{Agustsson_2017_CVPR_Workshops} and test on the given input images. However, they are not designed for large input as the practical scenario of SR usually needs to super-resolve large input like 2K to higher resolution (4K/8K).

{\bf Single Image Super-Resolution with Large Input} With the development of display devices, the resolutions of monitors have reached 4K or even 8K. Recently, there are some methods of super-resolving large input to a higher resolution. Since the computational and memory cost grows quadratically with the spatial resolution, recent methods all split the large input into local patches and merge the SR patches to the output. ClassSR \cite{kong2021classsr} is the seminal work that explores the problem of SISR with large input. They use a classification network to choose the restoration difficulty of each patch and allocate the optimal subnet for SR. However, ClassSR has two limitations. Firstly, it requires storing all the subnets. Secondly, the classification network brings additional computational costs. To solve the above limitations, APE \cite{wang2022adaptive} proposes to train a supernet for weight sharing. They use a regressor to predict the incremental capacity of each for the patch. ARM \cite{chen2022arm} also trains a supernet for weight sharing, however their subnets are constructed by reducing the channels instead of layers. CADyQ \cite{hong2022cadyq} uses network quantization to design the content-aware subnets. They train an MLP bit selector to determine the proper bit for each layer based on the content of an input patch. However, the MLP bit selector brings additional computational and storage costs. Besides, the selected bit is not optimal due to the uniform sampling.

{\bf Network Quantization} Network quantization is a very effective technique to accelerate the speed. They map the 32-bit floating point values of feature and weight to lower bit values \cite{lee2021network,jung2019learning,esser2019learned,choi2018pact,cai2017deep,zhou2017incremental}. Recent works also propose to allocate different bit-widths to different layers \cite{dong2019hawq,dong2020hawq,yao2021hawq}. However, these works are mainly focused on high-level visual understanding tasks such as image classification. Different from high-level tasks, super-resolution is more sensitive to the network quantization. PAMS \cite{Li2020PAMSQS} proposes to train the learnable upper bounds for quantization due to the absensce of BN layers. DAQ \cite{hong2022daq} uses different quantization parameters for each feature channel. DDTB \cite{zhong2022dynamic} presents a novel activation quantizer to accommodate the asymmetry of the activations. CADyQ \cite{hong2022cadyq} designs mixed precision subnets for the input patch and uses MLP selectors to determine the bit configuration. However, the introduced MLP brings additional computational and storage cost to SR networks.

\section{Method}

\subsection{Preliminary}
\label{subsec:preliminary}
In this section, we first introduce the background of network quantization since our method uses quantization to construct the subnets. For a given input activation $x$, the quantized output $x_{q}$ can be formulated as:

\begin{equation} \label{quantization function}
	\begin{aligned}
		x_{q} = \lfloor \frac{clamp(x)}{s(n)} \rceil s(n),
	\end{aligned}
\end{equation}
where $clamp(x)=max(min(x,\alpha),-\alpha)$ is the clamping function that uses a trainable upper bound to limit the range of input and $s(n)=\frac{\alpha}{2^{n-1}-1}$ is the mapping function that symmetrically scales the input to low bit output. The quantization of weights is similar to activations. Different from activations, we use fixed bit-width to quantize weights following existing methods \cite{Li2020PAMSQS,hong2022cadyq}. For a quantized model, the complexity can be measured by the number of operations weighted by the bit-widths (BitOPs).

Our CABM method builds a supernet $\mathcal{C}{_{W^*}^{T}}$ for SISR tasks, where $W^*$ denotes the quantized weight and $T$ denotes the activation bit configurations obtained by the Edge-to-Bit lookup tables. All the subnets share the same weight of supernet and each subnet can be represented as $\mathcal{C}{_{W^*}^{t_{s(p)}}}$, where $t_{s(p)}$ represents a certain bit configuration given patch $p$ and $s(\cdot)$ determines which subinterval the edge score of $p$ belongs to.

\subsection{Motivation}
\label{sec:motivation}
\begin{figure}[t!]
	\centering
	\includegraphics[height=7.5cm]{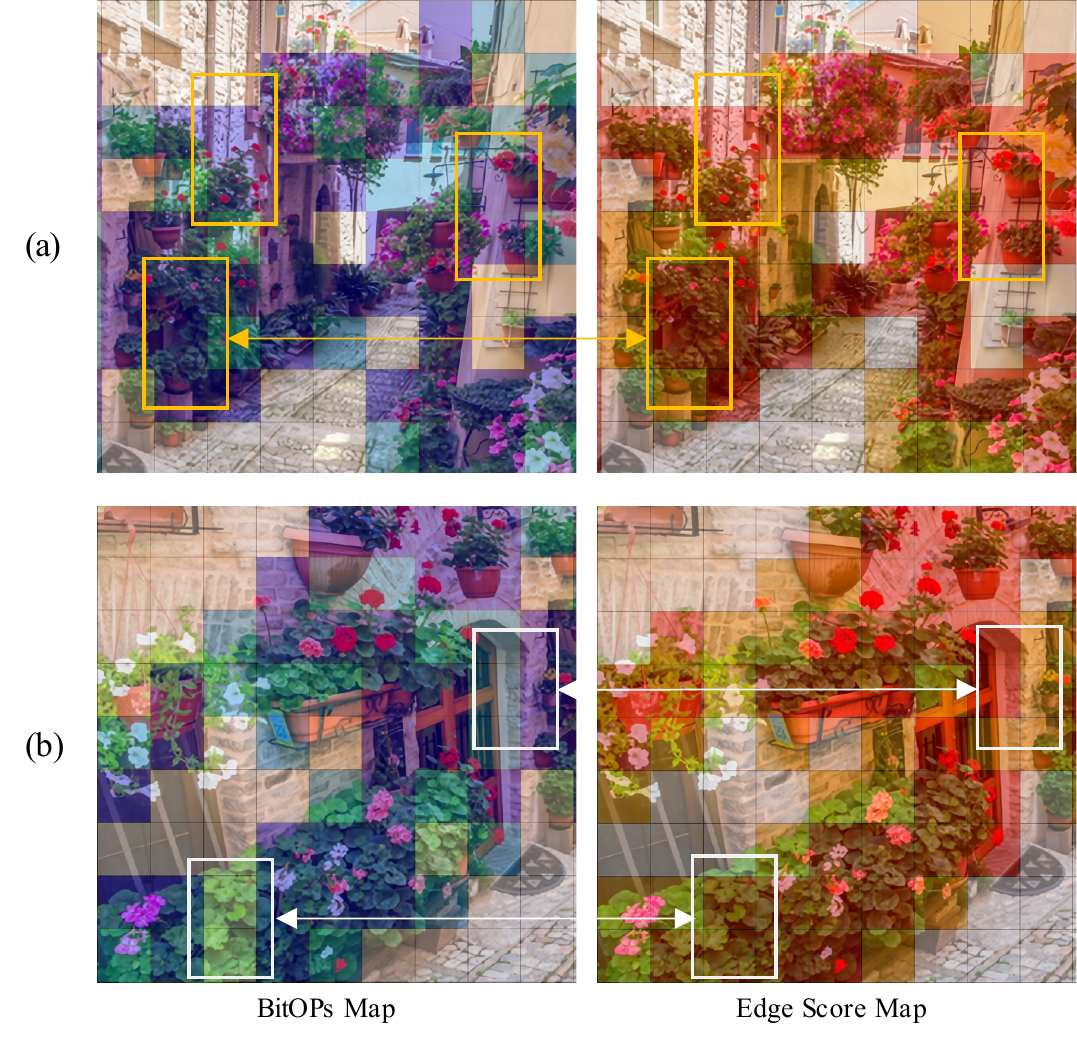}
	\caption{The motivation of our method. The darker color in the image means higher BitOPs / Edge Score. As can be seen in (a), patches with high edge scores often correspond to the bit configurations with high BitOPs. However, (b) shows that they are not always positively correlated, indicating the MLP selectors might fail to find optimal bit configurations.}
	\label{fig:flowers}
	\vskip -0.2cm
\end{figure}

Recent works \cite{hong2022cadyq, Liu2022InstanceAwareDN} have introduced additional modules such as MLP to adaptively determine the network quantization. Based on the results of MLPs, we find that MLP selectors usually choose bit configurations with high BitOPs for those patches with high edge scores. However, we notice that MLP selectors sometimes choose low BitOPs for those patches with high edge scores as shown in Fig.~\ref{fig:flowers}. Based on this observation, we realize that the bit configuration determined by MLP selectors might not be optimal. This is because recent methods uniformly sample the subnets for training, which makes simple subnets tend to overfit the input while complicated subnets tend to underfit the input.

\begin{table}[t]
	\centering
	\caption{The comparison of uniform sampling and BitOPs sampling. Feature Average Bit (FAB), PSNR, SSIM are reported for EDSR on Urban100. U and B denote uniform sampling and BitOPs sampling respectively.}
	\label{tab:motivationcomparison}
	\resizebox{.7\linewidth}{!}
	{
		\begin{tabular}{c|c|c|c} 
			\toprule[1.2pt]
			Model & FAB & PSNR & SSIM \\ 
			\hline
			EDSR-U & 6.20 & 25.94 & 0.782 \\
			EDSR-B & 6.20 & 26.01 & ~0.783 \vspace{-0.2em} \\
			\bottomrule[1.2pt]
		\end{tabular}
	}
	\vspace{-1.2em}
\end{table}
To further demonstrate the problem of uniform sampling, we conduct an experiment that samples the subnets according to the BitOPs. We define three levels of difficulties for subnets: easy, medium, and hard. The probability of sampling each type is calculated as follows.
\begin{equation}\label{probability}
	l^m = (\frac{N_m \cdot {\sum{_{k=1}^{N_m}}{BitOPs^2(\mathcal{C}{_{W^*}^{t_{s(k)}}})} }}{\sum_{m=1}^3(N_m \cdot \sum{_{k=1}^{N_m}}{BitOPs^2(\mathcal{C}{_{W^*}^{t_{s(k)}}})})}),
\end{equation}
where $l^m$ denotes the probability of choosing $m^{th}$ level, $BitOPs (\cdot)$ calculates the BitOPs of a subnet, $N_m$ indicates the number of samples belongs to each level. As shown in Tab.~\ref{tab:motivationcomparison}, sampling based on BitOPs achieves better performance compared with uniform sampling. Therefore, we believe the bit configuration determined by MLP selectors is not optimal. In this paper, we propose a novel method named Content-Aware Bit Mapping (CABM) to choose the optimal bit configuration for SISR networks.

\subsection{CABM Supernet Training}
\label{subsec:CABMtraining}
{\bf Train a Supernet with MLP Selectors} We first need to train a supernet that can adaptively generate all the bit configurations for model inference. Following existing methods \cite{hong2022cadyq, Liu2022InstanceAwareDN}, we introduce a supernet $\mathcal{M}{_{W}^{A}}$ that uses MLP to automatically decide bit configurations for various inputs. We choose the standard deviation of each layer's feature and the edge score as the input of each MLP selector. Specifically, given a training set $\{L, H\}$, where $L=\{l_n\}_{n=1...K}$ are the low-resolution (LR) images, $H=\{h_n\}_{n=1...K}$ are the high-resolution (HR) images. The training process of $\mathcal{M}{_{W}^{A}}$ is to balance the overall SR performance and the BitOPs of each subnet. Since this part is not our contribution, we simply summarize the process. For the details, we refer readers to the recent works \cite{hong2022cadyq, Liu2022InstanceAwareDN}.
\begin{figure}[t!]
	\centering
	\includegraphics[height=4cm]{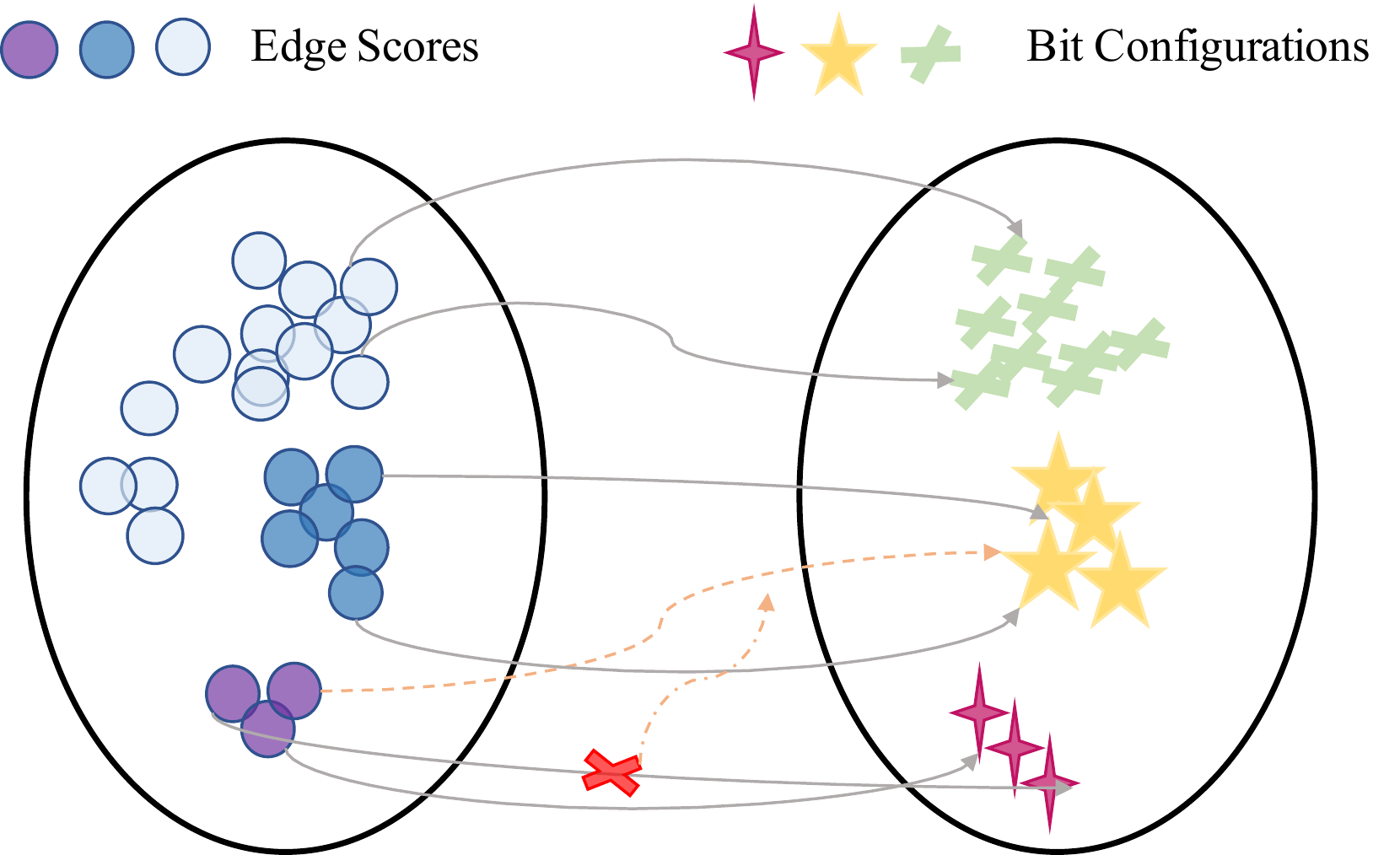}
	\vskip -0cm
	\caption{The illustration of Edge-to-Bit mappings.}
	\label{fig:mappings}
	\vskip -0.3cm
\end{figure}


\begin{table*}[t!]
	\centering
	\caption{Quantitative comparison of full precision models, PAMS, CADyQ and our method on Urban100, Test2K and Test4K. Computational complexity is measured by BitOPs of the backbone network for generating a 720p/2K/4K image accordingly. Feature Average Bit (FAB) and PSNR (dB) /SSIM results are also reported for each model. The scaling factor is x4.}
	\vspace{-1.0em}
	\label{tab:maincomparison_x4}
	\resizebox{.99\linewidth}{!}
	{
		\begin{tabular}{c|ccc|ccc|ccc} 
			\toprule[1.5pt]
			Model & BitOPs & FAB & Urban100 & BitOPs & FAB & Test2K & BitOPs & FAB & Test4K \\ 
			\hline
			\rule{0pt}{10pt}
			CARN  & 90.88G & 32.00 & 25.91/0.779 & 363.53G & 32.00 & 27.54/0.772 & 1.26T & 32.00 & 28.92/0.818  \\
			CARN-PAMS & 5.68G & 8.00 & 25.80/0.776  & 22.72G & 8.00 & 27.49/0.770  & 78.75G & 8.00 & 28.86/0.817  \\ 
			CARN-CADyQ & 3.13G & 4.99 & 25.88/0.778  & 10.55G & 4.40 & 27.52/0.770  & 38.11G & 4.36 & 28.87/0.817 \\
			\hline
			\rowcolor{teal!15}
			\rule{0pt}{10pt}
			CARN-CABM (Ours) & 1.57G & 4.21 & 25.90/0.779  & 6.11G & 4.15 & 27.51/0.771 & 20.89G & 4.12 & 28.88/0.817  \\
			\hline\hline
			\rule{0pt}{10pt}
			IDN & 81.88G & 32.00 & 25.46/0.764 & 327.52G & 32.00 & 27.40/0.766 & 1.13T & 32.00 & 28.73/0.813 \\
			IDN-PAMS & 5.12G & 8.00 & 25.56/0.768 & 20.47G & 8.00 & 27.43/0.767 & 70.62G & 8.00 & 28.77/0.814 \\ 
			IDN-CADyQ & 3.33G & 5.14 & 25.58/0.769 & 11.33G & 4.49 & 27.42/0.766 & 41.47G & 4.46 & 28.75/0.813 \\
			\hline
			\rowcolor{teal!15}
			\rule{0pt}{10pt}
			IDN-CABM (Ours) & 1.46G & 4.28 & 25.57/0.768 & 5.77G & 4.25 & 27.42/0.766 & 19.75G & 4.23 & 28.74/0.813 \\
			\hline\hline
			\rule{0pt}{10pt}
			SRResNet & 146.44G & 32.00 & 25.74/0.770 & 585.75G & 32.00 & 27.50/0.770 & 2.03T & 32.00 & 28.87/0.817 \\
			SRResNet-PAMS & 9.15G & 8.00 & 25.85/0.778 & 36.61G & 8.00 & 27.52/0.771 & 126.88G & 8.00 & 28.90/0.818 \\ 
			SRResNet-CADyQ  & 6.90G & 6.07 & 25.87/0.778 & 24.97G & 5.68 & 27.52/0.770 & 87.58G & 5.57 & 28.88/0.817 \\ 
			\hline
			\rowcolor{teal!15}
			\rule{0pt}{10pt}
			SRResNet-CABM (Ours) & 4.08G & 5.34 & 25.86/0.778 & 15.29G & 5.17 & 27.52/0.771 & 50.96G & 5.07 & 28.91/0.818 \\ 
			\hline\hline
			\rule{0pt}{10pt}
			EDSR & 114.23G & 32.00 & 26.03/0.784 & 456.92G & 32.00 & 27.59/0.773 & 1.58T & 32.00 & 28.99/0.820 \\
			EDSR-PAMS & 7.14G & 8.00 & 26.01/0.784 & 28.56G & 8.00 & 27.59/0.773 & 98.75G & 8.00 & 28.99/0.820 \\ 
			EDSR-CADyQ & 6.64G & 6.70 & 25.90/0.781 & 23.18G & 6.11 & 27.54/0.771 & 82.37G & 6.06 & 28.91/0.818 \\ 
			\hline
			\rowcolor{teal!15}
			\rule{0pt}{10pt}
			EDSR-CABM (Ours) & 3.75G & 5.80 & 25.95/0.782 & 14.24G & 5.65 & 27.57/0.772 & 47.70G & 5.56 & ~28.96/0.819 \vspace{-0.1cm} \\
			\bottomrule[1.5pt]
		\end{tabular}
	}
\end{table*}

\begin{table*}[t!]
	\centering
	\caption{Quantitative comparison of full precision models, PAMS, CADyQ and our method on Urban100, Test2K and Test4K. Computational complexity is measured by BitOPs of the backbone network for generating a 720p/2K/4K image accordingly. Feature Average Bit (FAB) and PSNR (dB) /SSIM results are also reported for each model. The scaling factor is x2.}
	\vspace{-1.0em}
	\label{tab:maincomparison_x2}
	\resizebox{.99\linewidth}{!}
	{
		\begin{tabular}{c|ccc|ccc|ccc} 
			\toprule[1.5pt]
			Model & BitOPs & FAB & Urban100 & BitOPs & FAB & Test2K & BitOPs & FAB & Test4K \\ 
			\hline
			\rule{0pt}{10pt}
			CARN  & 222.83G & 32.00 & 31.93/0.926 & 891.33G & 32.00 & 32.77/0.928 & 3.08T & 32.00 & 34.34/0.943  \\
			CARN-PAMS & 13.93G & 8.00 & 31.97/0.927  & 55.71G & 8.00 & 32.75/0.928  & 192.50G & 8.00 & 34.36/0.943  \\ 
			CARN-CADyQ & 5.25G & 4.46 & 31.82/0.925  & 19.18G & 4.22 & 32.69/0.927  & 67.53G & 4.19 & 34.29/0.942 \\
			\hline
			\rowcolor{teal!15}
			\rule{0pt}{10pt}
			CARN-CABM (Ours) & 3.61G & 4.09 & 31.96/0.927  & 14.35G & 4.06 & 32.75/0.928 & 49.58G & 4.06 & 34.35/0.923  \\
			\hline\hline
			\rule{0pt}{10pt}
			IDN & 174.10G & 32.00 & 31.22/0.919 & 696.41G & 32.00 & 32.39/0.923 & 2.41T & 32.00 & 34.00/0.940 \\
			IDN-PAMS & 10.88G & 8.00 & 31.28/0.920 & 43.53G & 8.00 & 32.46/0.925 & 151.25G & 8.00 & 34.04/0.941 \\ 
			IDN-CADyQ & 5.96G & 5.28 & 31.34/0.921 & 18.35G & 4.45 & 32.48/0.925 & 66.11G & 4.44 & 34.07/0.941 \\
			\hline
			\rowcolor{teal!15}
			\rule{0pt}{10pt}
			IDN-CABM (Ours) & 3.01G & 4.21 & 31.40/0.921 & 11.94G & 4.19 & 32.50/0.925 & 41.49G & 4.19 & 34.10/0.941 \\
			\hline\hline
			\rule{0pt}{10pt}
			SRResNet & 406.63G & 32.00 & 31.51/0.922 & 1.63T & 32.00 & 32.56/0.925 & 5.62T & 32.00 & 34.15/0.941 \\
			SRResNet-PAMS & 25.41G & 8.00 & 31.55/0.923 & 101.88G & 8.00 & 32.53/0.925 & 351.25G & 8.00 & 34.17/0.942 \\ 
			SRResNet-CADyQ  & 17.04G & 6.23 & 31.48/0.922 & 64.60G & 6.04 & 32.50/0.925 & 222.34G & 5.98 & 34.12/0.941 \\ 
			\hline
			\rowcolor{teal!15}
			\rule{0pt}{10pt}
			SRResNet-CABM (Ours) & 11.84G & 5.46 & 31.54/0.923 & 45.22G & 5.33 & 32.55/0.925 & 150.12G & 5.23 & 34.16/0.942 \\ 
			\hline\hline
			\rule{0pt}{10pt}
			EDSR & 316.25G & 32.00 & 31.97/0.927 & 1.27T & 32.00 & 32.75/0.928 & 4.37T & 32.00 & 34.37/0.943 \\
			EDSR-PAMS & 19.77G & 8.00 & 32.06/0.928 & 79.38G & 8.00 & 32.79/0.928 & 273.13G & 8.00 & 34.42/0.944 \\ 
			EDSR-CADyQ & 12.86G & 6.03 & 31.84/0.925 & 51.01G & 5.88 & 32.68/0.927 & 169.15G & 5.79 & 34.28/0.942 \\ 
			\hline
			\rowcolor{teal!15}
			\rule{0pt}{10pt}
			EDSR-CABM (Ours) & 9.65G & 5.59 & 31.92/0.927 & 36.03G & 5.39 & 32.74/0.927 & 120.33G & 5.31 & ~34.33/0.943 \vspace{-0.1cm} \\
			\bottomrule[1.5pt]
		\end{tabular}
	}
	\vspace{-0.6em}
\end{table*}

{\bf Build a Supernet with CABM} As illustrated in the motivation, the uniform sampling makes simple subnets overfitted and complicated subnets underfitted. Therefore, the bit configuration of MLP selectors is not optimal for the given input patch. MLP selectors use two kinds of information to decide the bit for each layer. Among them, the standard deviation represents the feature importance of the current layer while the edge score of a patch is constant through layers. For different patches, the importance of layers might be different. However, for patches with the same edge score, the layer difference is almost negligible. This inspires us to build Edge-to-Bit lookup tables for determining the optimal bit configuration. 

To be more specific, for all LR patches on the validation set of DIV2K, we compute the edge scores denoted as $E = \{e_i\}_{i=1...O}$. Then we use $\mathcal{M}{_{W}^{A}}$ to generate the corresponding bit configurations for all the patches. Assume that the edge detection retains precision $F$, \emph{e.g.} 0.01, we can then split the edge score interval $[0, max(E)]$ to $R=\frac{10 \cdot max(E) + F}{5 \cdot F}$ subintervals $S=\{s_r\}_{r=1...R}$ so that $r^{th}$ subinterval can be defined as:
\begin{equation}
	s_r = [ \frac{F \cdot (r-1)}{2}, \frac{F \cdot (5r-1)}{10}].
	\label{subinterval}
\end{equation}
Given a patch $\hat{p}$ with its edge score $D(\hat{p})$, we can determine the index of subinterval for $D(\hat{p})$ according to the $R$ subintervals. We denote the process that determines the subinterval of $\hat{p}$ as $s(\cdot)$. For those subintervals without corresponding bit configurations, we choose the bit configurations from the nearest subintervals for them. Therefore, we can build a simple mapping between edge scores and bit configurations using $\mathcal{M}{_{W}^{A}}$.
\begin{equation}
	s(D(\widehat p)) \in S \to \overline T  = {\{ \overline {{t_i}} \} _{i = 1...O}}
\end{equation}

However, one subinterval might correspond to thousands of bit configurations. We design a simple yet effective strategy to determine one optimal bit configuration for each subinterval. We observe that high bit widths retain more information of features and improve the accuracy of quantized models. However, we need to minimize the computational cost using low bit widths. Therefore, we first sample the bit configuration with minimum BitOPs for each subinterval. However, for a small number of patches within a certain precision range, the bit configurations determined by MLPs may fall into the local minimum. Our solution to this problem is expanding the range of subintervals. Thus each expanded subinterval can contain more bit configurations. The adjusted subintervals can be formulated as:
\begin{equation}
	s_r = \left\{\begin{array}{ll}{[ \frac{F \cdot (r-1)}{2}, \frac{F \cdot (5r-1)}{10}],} & {r \in[0, \beta]}, \\ {[ \frac{F \cdot (r-1)}{2} - \Delta e, \frac{F \cdot (5r-1)}{10} + \Delta e],} & {r \in (\beta, R]}. \end{array}\right.
	\label{subinterval2}
\end{equation}
where ${\Delta e}$ and ${\beta}$ are hyper-parameters.
In this way, we can find one optimal bit configuration by choosing the bit configuration with minimum BitOPs for each expanded subinterval. We build the optimal one-to-one Edge-to-Bit lookup tables $ T = \{t_r\}_{r=1...R}$ to reduce the BitOPs of MLP selectors. The process of CABM is illustrated by Fig.~\ref{fig:mappings}. After building the Edge-to-Bit lookup tables with CABM to construct the supernet $\mathcal{C}{_{W^*}^{T}}$, we conduct a simple fine-tuning for the CABM supernet: 

\begin{equation}\label{LUT optimize}
	\min \sum_{n=1}^B \sum_{p=1}^{P_n}\| \mathcal{C}{_{W^*}^{t_{s(p)}}}(l{_{p}^n}) - h{_{p}^n} \|_1,
\end{equation}
where $\|\cdot\|_1$ denotes the commonly-used $\ell_1$-norm in the SISR task, $l_{p} \in l_{n} = \{l_p\}_{p=1...P} \in \{l_n\}_{n=1...B}$ and $h_{p} \in h_{n} = \{h_p\}_{p=1...P} \in \{h_n\}_{n=1...B}$ are respectively the LR image patch and HR image patch in the current training iteration, $B$ is the batch size and $P_n$ is the patch number for each image. $W^*$ is initialized from the weight $W$ of $\mathcal{M}{_{W}^{A}}$.

\begin{figure*}[t]
	\centering
	\includegraphics[width=0.92\textwidth]{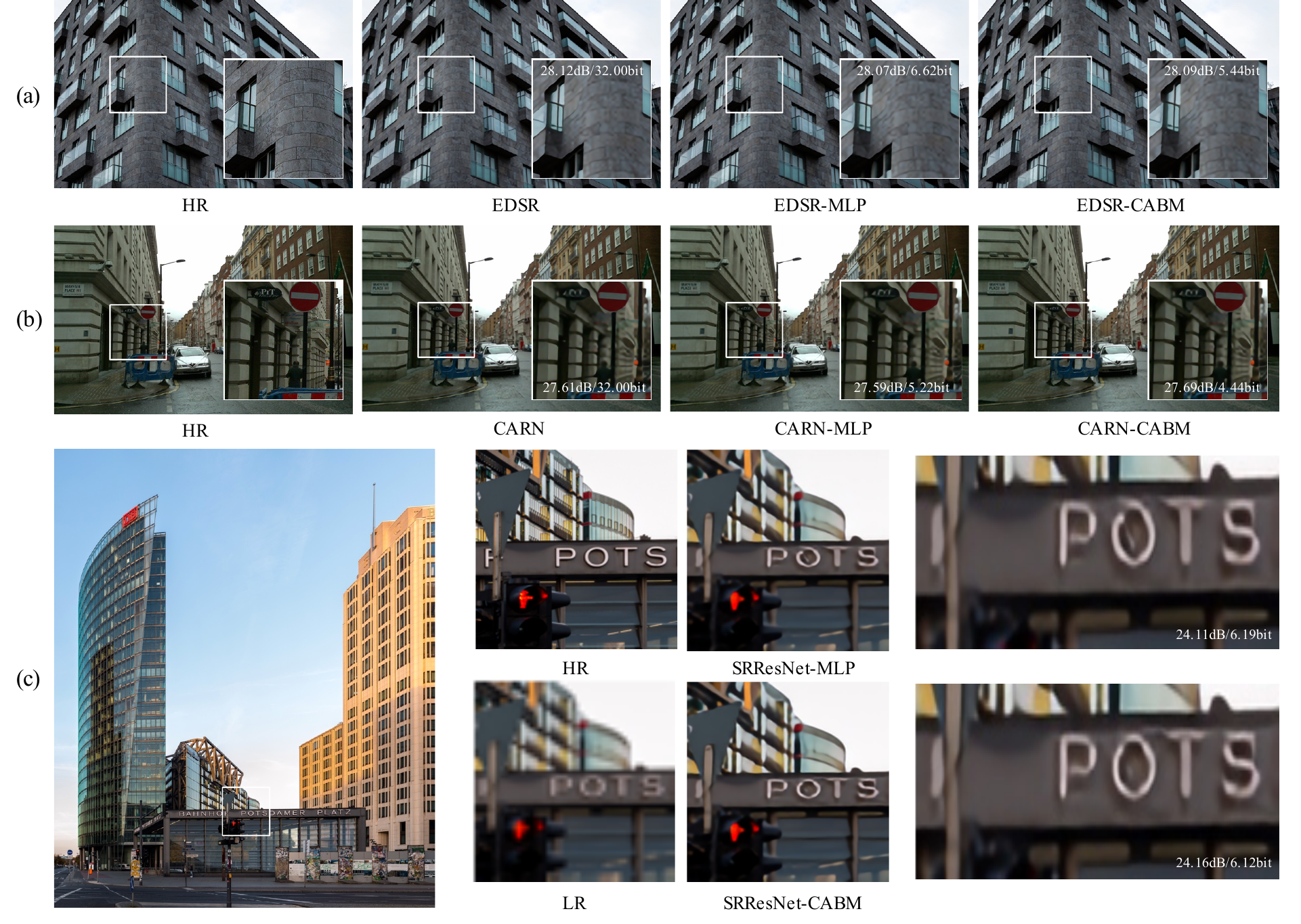}
	\caption{(a) (b) Qualitative comparison on two images from Urban100 with EDSR/CARN. (c) (d) Qualitative comparison on one image from Test4K with SRResNet. As can be seen, our method can achieve similar performance while reduce the FAB.}
	\label{fig:visual_cmp}
\end{figure*}

\subsection{CABM Supernet Inference}
During inference, we first split the whole large input into local patches of a given size, and then use the Laplacian edge detection operator to calculate the edge scores. Based on the edge scores, we can quickly obtain the corresponding subnet for each patch from CABM supernet. After super-resolving all the patches, we merge the SR patches to the output. Compared with MLP selectors, CABM can achieve lower BitOPs with negligible additional computational cost. The whole pipeline of CABM is shown in Fig.~\ref{fig:pipeline}.

\begin{figure*}[t]
	\centering
	\includegraphics[width=0.95\textwidth]{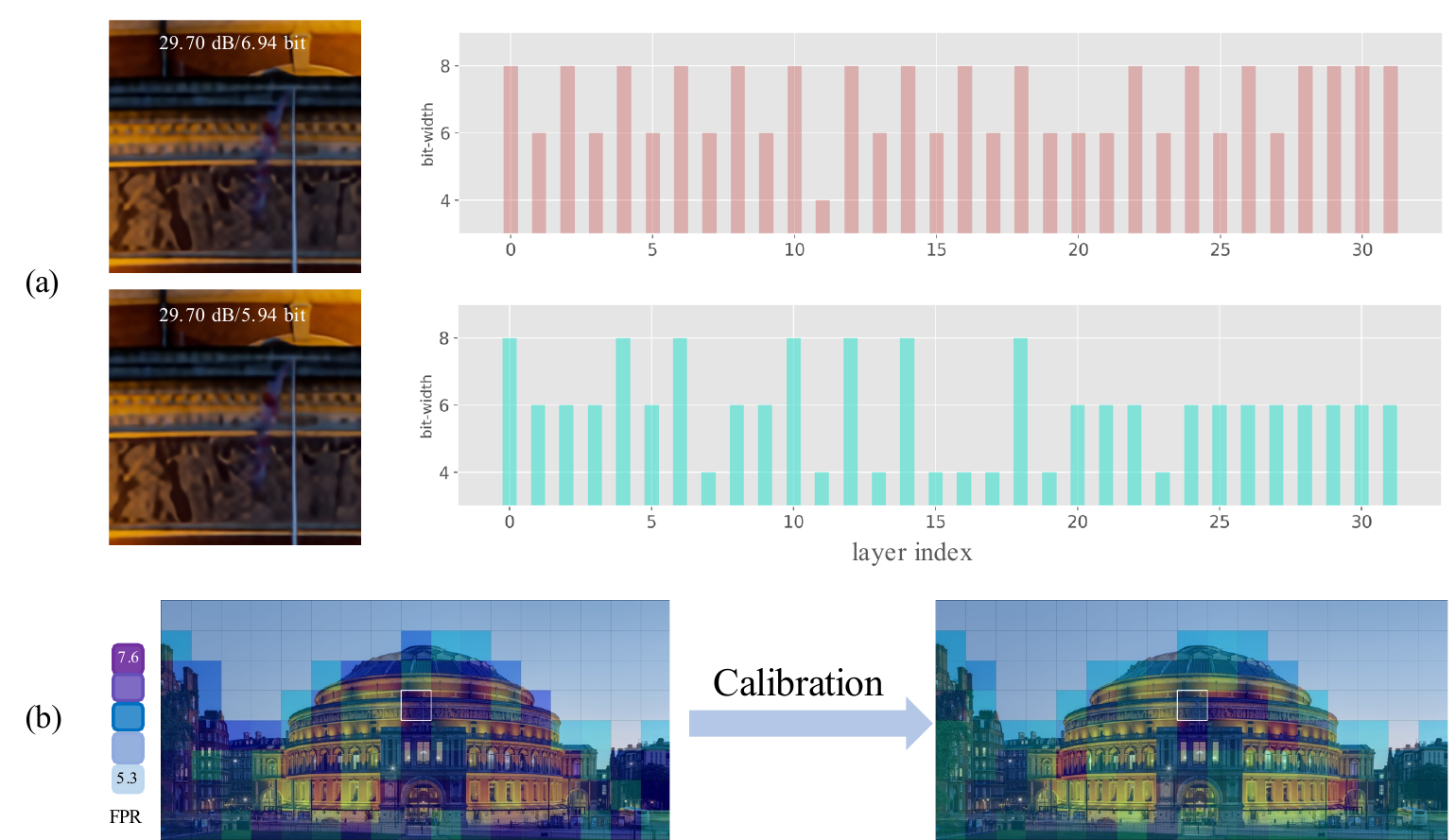}
	\caption{Quantitative comparison on an image in Test4K with backbone network EDSR. (a) Comparison between patches before and after calibration, and the corresponding bit configuration of different layers. (b) FAB heat maps before and after calibration.}
	\label{fig:bit_cmp}
\end{figure*}
\section{Experiment}
\subsection{Experimental Settings}
\label{subsec:expsettings}
To verify the generalization and effectiveness of our method, we conduct detailed experiments on four representative models. For plain models, we choose two widely used SISR models, \emph{i.e.} EDSR \cite{lim2017enhanced} and SRResNet \cite{ledig2017photo}. Since the quantization for activations mainly affects the flow of information between blocks, it is not enough to apply our CABM to plain models. Therefore, we apply CABM to IDN \cite{Hui2018FastAA} and CARN \cite{Ahn2018FastAA} that have a hierarchical feature extraction mechanism. Besides, we notice that the leaning-based quantization method we choose performs better when bit-width is greater than 4, and Tensor Cores mainly support 4/8-bit, so it is reasonable to select 4/6/8-bit as our candidate bits. As for quantization details of activation and weight, we follow the settings of previous works \cite{hong2022cadyq, Zhong2022DynamicDT}.

{\bf Implementation Details} 
All models are trained on DIV2K datasets \cite{Agustsson_2017_CVPR_Workshops} which contains 800 images for training, 100 images for validation, and 100 images for testing. The lookup tables are built based on the validation set using the proposed CABM method and the precision of edge scores is $F=0.01$. For CABM fine-tuning, we use the pre-trained weight from supernet $\mathcal{M}{_{W}^{A}}$. For testing, we use the Peak Signal-to-Noise Ratio (PSNR) and Structural Similarity (SSIM) as the metrics to evaluate the SR performance of all methods on three test datasets: benchmark \cite{Huang2015SingleIS}, Test2K and Test4K \cite{kong2021classsr}. In Test2k and Test 4k, images are generated following previous work from DIV8K datasets (index 1201-1400) \cite{Gu2019DIV8KD8}. Without special mention, all test input images are split to 96$\times$96 LR patches with scaling factor x4. For EDSR and SRResNet, ${\Delta e}$ and ${\beta}$ are respectively set to 10 and 9000. For CARN and IDN, they are set to 10 and 6000.
\subsection{Quantitative and Qualitative Results}
To fully prove the effectiveness and generalization of our proposed CABM method, we compare our results with full precision models, PAMS \cite{Li2020PAMSQS} and CADyQ \cite{hong2022cadyq}. PAMS is a SISR quantization method, which uses fixed bit-width for network. CADyQ uses MLPs to adaptively adjust the bit-widths according to the input patch similar to our method. All the models are trained by ourselves using the official codebases and instructions to avoid unfair comparison. 
As shown in Tab.~\ref{tab:maincomparison_x4}, our CABM method reduces the computational overhead and achieves the accuracy as full precision models on scaling factor x4. For SRResNet, CABM outperforms the full precision model by 0.11 dB (PSNR) and 0.008 (SSIM) with only 2.8\% BitOPs. As for IDN, it is obvious that MLP bit selectors fail to learn the optimal bit configurations for the quantized models. Besides, the introduced MLPs brings additional computational cost to SR networks. Compared with them, our CABM is much more efficient when the BitOPs of quantized models are low enough since the lookup tables bring negligible additional computational cost. We also show the comparison on scaling factor x2 in Tab.~\ref{tab:maincomparison_x2}. Although our method fails to achieve competitive results compared with full precision models, our method obtains similar results as PAMS and CADyQ while our FAB is much lower. Fig.~\ref{fig:visual_cmp} shows the qualitative results. To summarize, our CABM method achieves almost lossless model performance compared with PAMS and CADyQ while significantly reduces the computational overhead.

\begin{table}[t]
	\centering
	\caption{Ablation study of CABM Fine-tuning. FAB, PSNR (dB), and SSIM results are reported for each model on Set14 and Urban100 datasets.}
	\label{tab:ablation0}
	\resizebox{.95\linewidth}{!}
	{
		\begin{tabular}{c|cc|cc} 
			\toprule[1.5pt]
			Datasets &\multicolumn{2}{c|}{Set14} & \multicolumn{2}{c}{Urban100} \\
			\hline
			Model & FAB & PSNR/SSIM & FAB & PSNR/SSIM \\ 
			\hline
			\rule{0pt}{10pt}
			EDSR-$\mathcal{M}{_{W}^{A}}$ & 6.37 & 28.53/0.780 & 6.70 & 25.90/0.781 \\
			\rule{0pt}{10pt}
			EDSR-$\mathcal{C}{_{W}^{T}}$ & 5.80 & 28.47/0.777 & 5.80 & 25.77/0.774 \\
			\rule{0pt}{10pt}
			IDN-$\mathcal{M}{_{W}^{A}}$ & 4.72 & 28.34/0.774 & 5.14 & 25.58/0.769 \\
			\rule{0pt}{10pt}
			IDN-$\mathcal{C}{_{W}^{T}}$ & 4.18 & 27.08/0.750 & 4.28 & 24.32/0.732 \\
			\bottomrule[1.5pt]
		\end{tabular}
	}
	\vspace{-1.2em}
\end{table}

\subsection{Ablation Study}
{\bf CABM Fine-tuning} 
After we get the MLP supernet $\mathcal{M}{_{W}^{A}}$ and generate our CABM look-up tables, one straightforward choice is using the original weight $W$ to build a CABM supernet $\mathcal{C}{_{W}^{T}}$. Instead, the proposed method conduct the fine-tuning after CABM. To verify the effectiveness of CABM fine-tuning, we show the results in Tab.~\ref{tab:ablation0}. The experimental settings are consistent with our main results. Without fine-tuning, $\mathcal{C}{_{W}^{T}}$ fails to achieve reasonable results after removing MLPs. Especially for IDN \cite{Hui2018FastAA}, a network with dense information, the PSNR value is 1.26 dB lower. This further provides experimental support to our motivation in Sec.~\ref{sec:motivation} that bit configurations and edge scores are not always positively correlated. Therefore, in order to adapt each layer, fine-tuning is necessary after CABM.

\begin{table}[t]
	\centering
	\caption{The comparison of different calibration settings on EDSR with CABM. FAB, PSNR (dB), and SSIM results are reported for each setting on Urban100 datasets.}
	\label{tab:ablation1}
	\resizebox{.9\linewidth}{!}
	{
		\begin{tabular}{c|c|cc|cc} 
			\toprule[1.5pt]
			& &\multicolumn{2}{c|}{Uniform} & \multicolumn{2}{c}{BitOPs} \\
			\hline
			${\Delta e}$ & FAB & PSNR & SSIM & PSNR & SSIM \\ 
			\hline
			10 & 5.80 & 25.91 & 0.781 & 25.95 & 0.782 \\
			20 & 5.71 & 25.91 & 0.781 & 25.93 & 0.781 \\
			30 & 5.65 & 25.90 & 0.781 & 25.91 & 0.780 \\ 
			40 & 5.60 & 25.90 & 0.780 & 25.89 & 0.779 \\
			80 & 5.54 & 25.90 & 0.780 & 25.87 & 0.778 \\
			\bottomrule[1.5pt]
		\end{tabular}
	}
	\vspace{-1.2em}
\end{table}
\begin{table}[t]
	\centering
	\caption{The comparison of different strategies for bit selecting on EDSR with CABM. FAB, PSNR (dB), and SSIM results are reported for each model on Set14 and Urban100 datasets.}
	\label{tab:ablation2}
	\resizebox{.99\linewidth}{!}
	{
		\begin{tabular}{c|cc|cc} 
			\toprule[1.5pt]
			Datasets &\multicolumn{2}{c|}{Set14} & \multicolumn{2}{c}{Urban100} \\
			\hline
			\rule{0pt}{10pt}
			Model & FAB & PSNR/SSIM & FAB & PSNR/SSIM \\ 
			\hline
			\rule{0pt}{10pt}
			EDSR-baseline & 32.00 & 28.58/0.781 & 32.00 & 26.03/0.784 \\
			\rule{0pt}{10pt}
			$\mathcal{C}{_{W^*}^{T1}}$ & 5.92 & 28.56/0.780 & 5.90 & 25.95/0.782 \\
			\rule{0pt}{10pt}
			$\mathcal{C}{_{W^*}^{T2}}$ & 6.69 & 28.58/0.781 & 6.64 & 26.04/0.784 \\
			\rule{0pt}{10pt}
			$\mathcal{C}{_{W^*}^{T3}}$ & 6.21 & 28.57/0.781 & 6.20 & 26.01/0.783 \\
			\bottomrule[1.5pt]
		\end{tabular}
	}
	\vspace{-1.0em}
\end{table}

{\bf Different Calibration Settings} 
To evaluate our proposed CABM method in Sec.~\ref{subsec:CABMtraining}, we conduct more experiments to evaluate the impact of different settings. Specifically, we choose EDSR as our backbone and the $\beta$ is set to 9000 with edge precision $F=0.01$. Tab.~\ref{tab:ablation1} reports the quantitative results of different expanding settings. As we have mentioned, the results obtained by MLP selectors are often not optimal due to the uniform subnet sampling. As we increase the expanding range $\Delta e$, the possibilities of finding better solutions and worse solutions increase at the same time. By analyzing the results of expanding ranges, we can empirically set the optimal $\Delta e$ for the proposed CABM method. In Fig.~\ref{fig:bit_cmp}, we have compared the visual results of the mappings before and after our strategy, and $\Delta e$ is set to 80 here. As can be seen, both mappings can achieve similar performance, while our strategy can calibrate the mapping and further reduce 1 FAB. This demonstrate that our calibration method can improve the original one-to-many mapping and reduce the computational cost.

{\bf Bit Selection Strategies} Our method expands the range of subintervals and select one bit configuration with minimum BitOPs for each subinterval. We denote our minimum sampling strategy as S1. The other two alternatives are sampling one bit configuration with maximum BitOPs (S2) and randomly sampling one bit configuration (S3). To evaluate these bit selection strategies, we analyze and compare S1, S2 and S3. We fine-tune different supernets $\mathcal{C}{_{W^*}^{T}}$ and denote them as $\mathcal{C}{_{W^*}^{T1}}$, $\mathcal{C}{_{W^*}^{T2}}$ and $\mathcal{C}{_{W^*}^{T3}}$, where $T1$, $T2$ and $T3$ are different look-up tables generated by S1, S2, and S3 respectively. It can be observed from Tab.~\ref{tab:ablation2} that $\mathcal{C}{_{W^*}^{T2}}$ achieves the best performance. However, from the perspective of balancing performance and computation, $\mathcal{C}{_{W^*}^{T2}}$ may not be an appropriate choice since the PSNR on Set14 is only 0.02 dB higher than $\mathcal{C}{_{W^*}^{T1}}$ while the FAB is 0.77 higher. Therefore, we choose the S1 strategy for the proposed CABM.
\section{Limitation}
Although CABM can further reduce the computational overhead of existing SISR methods, our method still has some limitations. For example, our network uses mixed precision quantization, which requires specific hardware support to achieve practical speedup. Besides, applying CABM to Video Super-Resolution (VSR) is not a trivial task since VSR needs to consider the temporal information and has more modules compared with SISR. Mixed precision quantization of VSR networks is still unexplored to the best of our knowledge. These problems will be our future works.
\section{Conclusion}
To summarize, we propose a novel Content-Aware Bit Mapping (CABM) method for SISR with large input. Existing methods learn the MLP selectors to determine the bit configuration for a given patch. However, they uniformly sample the subnets and fail to obtain optimal bit configuration. Instead of using the MLPs, CABM builds the Edge-to-Bit lookup tables to determine the bit configuration. In order to further reduce the computational cost, we present a novel calibration strategy to find a better mapping between edge score and bit configuration. Our method can achieve similar performance as existing methods with negligible additional computational and storage cost.

{\small
\bibliographystyle{ieee_fullname}
\bibliography{PaperForReview}
}

\end{document}